# Human Languages with Greater Information Density Increase Communication Speed, but Decrease Conversation Breadth

**Authors:** Pedro Aceves[1], James A. Evans[2,3]*

**Affiliations:**

[1] Department of Management and Organization, Johns Hopkins University, Baltimore, MD 21202

[2] Department of Sociology & Knowledge Lab, University of Chicago, 5735 South Ellis Avenue, Chicago, IL 60637.

[3] Santa Fe Institute, 1399 Hyde Park Rd, Santa Fe, NM 87501

*Corresponding author. E-mail: jevans@uchicago.edu (J.E.)

**Human languages vary widely in how they encode information within circumscribed semantic domains (e.g., time, space, color, human body parts and activities), but little is known about the global structure of semantic information and nothing about its relation to human communication. We first show that across a sample of ~1,000 languages, there is broad variation in how densely languages encode information into their words. Second, we show that this language information density is associated with a denser configuration of semantic information. Finally, we trace the relationship between language information density and patterns of communication, showing that informationally denser languages tend toward (1) faster communication, but (2) conceptually narrower conversations within which topics of conversation are discussed at greater depth. These results highlight an important source of variation across the human communicative channel, revealing that the structure of language shapes the nature and texture of human engagement, with consequences for human behavior across levels of society.**

Language is the primary medium through which human information is communicated and coordination is achieved[1–4]. One of the most important language functions is to categorize the world so messages can be communicated through conversation. While we know a great deal about how human languages vary in their encoding of information within circumscribed semantic domains[5,6] such as color[7–10], sound[11], number[12], locomotion[13,14], time[15–17], space[18–21], human activities[22,23], gender[24], body parts[25] and biology[26,27], little is known about the global structure of semantic information[28] and its effect on human communication[29].

The words of a language carve the vast space of meaning into concepts for communication. When we hear a sentence such as "I play soccer, monopoly, and the violin,"[30] the words within may strike the English speaker as natural representations of their underlying conceptual information and related associations. This traditional, universalist perspective[5] posits that conceptual information is represented by language through a straightforward mapping between words and underlying concepts, invariant across cultures[31–34]. Yet, translating this sentence into Spanish leads to the following sentence: "Yo juego fútbol y monopoly, y toco el violín" (in English "I play soccer and monopoly, and I *touch* the violin"). Within this circumscribed example, the word "play" represents more conceptual information in English than in Spanish by also containing the act of violin playing, suggesting the variability of conceptual information across languages (see Fig. 1a). This alternative, non-universalist perspective posits that cultures differ in how they carve experience into concepts and that languages vary in their correspondence between words and conceptual information[35–37]. Despite documented evidence of this variation across circumscribed knowledge domains using small language samples[7,9–27,34,38,39], and even recent larger-scale analyses that broaden the domains, number of languages or both[5,40], analyses of linguistic relativity have tended to ignore the degree to which the words of one language encode more or less conceptual information than another. Greater information density entails more polysemy, where words reference more meanings across the language. We know little about cross-linguistic information density and even less about its relation to human communication and knowledge creation.

We here begin to fill this lacuna by providing large-scale evidence to document wide variation in the degree of information density across the world's languages (step1 in Table 1). Information density refers to the average amount of conceptual information per language unit,[38,39] a notion closely related to other information theoretic measures of language such as entropy and efficiency[41,42]. To systematically estimate the information density of languages, we used 18 diverse parallel translation corpora that contain equivalent text in English and many other languages (see Methods). As the purpose of translation is to preserve the information, force, and meaning of an expression in a code appropriate to the target language and culture[43], we compare how pairs of languages differ in their linguistic encoding of the same conceptual information. We estimate information density by using the Huffman coding algorithm[44], an information-theoretic measure that translates word symbols from each language into their most efficient binary code given the symbol distribution within the document.

The Huffman algorithm allows us to translate each document from the parallel translation into a string of 0's and 1's that efficiently encodes each word, character, or other symbol based on its frequency or probability of recurrence. Coding common symbols with fewer bits (0's and 1's) than rare symbols minimizes the overall code length, which enables comparison of different language codes (e.g., English and Spanish) for a given translation. Languages that encode fewer bits are more informationally dense in that each bit represents more conceptual information. Because every corpus contains distinct information content, and every document exists in English as well as one or more other languages, we use the bit size ratio of the secondary language to English as a normalized information density measure to enable comparison across all corpora. For ease of interpretation, we multiply the ratio by -100, such that larger values indicate greater informational density. We create this language information density measure using data from 18 diverse corpora (14 billion tokens total) representing 998 languages from 101 language families and covering knowledge domains as diverse as medicine, technology, economics, politics, law, and entertainment.

Next we ask whether the information density of a language is associated with its density of the conceptual, semantic space (step 2 in Table 1). Conceptual space is the multidimensional space characterizing the distance between word and concept meanings in a language. The distribution of concepts within this space is what allows us to say, for example, that the "play" concept is closer to the "violin" concept in English than in Spanish. Following the distribution hypothesis, which posits that word meanings are proximate as a function of shared context[45], we estimate conceptual density with neural word embedding models built according to this principle[46]. Word embedding models have previously been demonstrated to represent rich semantic associations with geometric distances, reproducing human analogies[47,48], cultural associations[49] and bias[50,51], and revealing polysemy or multiple use[52,53]. Word embedding algorithms use word co-occurrences within a document and a neural network architecture to train a high-dimensional vector for each word in the corpus (see Methods). The output of the algorithm is a vector space wherein words with similar syntactic uses and semantic meaning tend to be close together in the space and words can have multiple dimensions of similarity and difference. The expectation is that languages with informationally more dense words will tend toward conceptual spaces that are likewise more dense, where the normalized distance between each word within the space is smaller, such that the average associational possibilities and polysemous uses for each word is greater[52]. We calculate this semantic density on the same document corpora for which we calculated information density. We note that it is not a foregone conclusion that information and conceptual density will correlate. Information density relies only on the frequency of each word across the corpus; Conceptual density relies on the diversity of contexts in which each word resides. If they strongly correlate, it would mean that the more times a word is used, the more ways it is used.

We then inquire whether representing the same conceptual information in fewer bits, as informationally dense languages do, sends information more quickly through spoken communication (step 3). The information theoretic expectation is that this should be the case, as fewer bits should be able to flow more quickly through a fixed bandwidth channel[54]. In the same way that an electronic document will be downloaded more quickly if it contains fewer bits, languages that encode the same information more densely (i.e., in fewer bits) should be able to communicate that information more quickly. To test this expectation, we measured the time it took to speak the audio Bible in 265 languages—that is, to communicate the same message encoded in different symbolic systems—and relate this to the information density measured in step 1 above.

We further consider whether differences in conceptual information density shape not only the speed of communication, but also discursive and conversational patterns in real-world communication, focusing primarily on the conceptual breadth of information brought to bear on a conversation (step 4). Recent work conceptualizes conversations as probabilistic random walks through the topic space of conversations[53,55,56]. At every point in a conversation there is a micro-topic of discussion, and each word in a language is in the set of words that could be used to talk about any given topic. Therefore, each potential topic has a probability distribution over the words that could be used within that topic. This is consistent with a more micro, cognitive view, where research on long-term memory retrieval[57] is carried out through verbal fluency tasks, finding that sets of semantically similar words are generated together[58,59] through a random-walk, associative retrieval process.[60,61] This means that words function as cues to activate associations in memory. Each word or subsequent set of words activates a new set of potential words from which to proceed. Consistent with optimal foraging theory,[62,63] the process of associative semantic search is carried out by exploiting local word patches until cues are depleted[57,64] and the individual undertakes global exploration in search of new word patches. The process that occurs during conversation is an extension of this process, but instead of internal word cues, the cues come from the utterances of others. As each new word is spoken, it serves as a cue that activates new traces in the memory of all participants, opening new possibilities for movement through the conceptual space during the conversation. We therefore expect that informationally denser languages will exhibit conversations that circle and retrace over narrower micro-topics, given that informationally denser words will facilitate the deeper discussion of any given topic by bringing more information to bear on that topic.

We traced how individuals traversed the conceptual space of a language in conversation, using text from over 5,800 conversations in 14 languages to measure (1) the total conceptual breadth of the conversation, and (2) the average conceptual distance traveled in each conversational turn. We measured the conceptual breadth of a conversation in multiple ways using embedding models. First, we measured the centroid vector of all words used during a conversation. We then measured how far away, on average, each unique word was to this centroid vector. This leaves us with a measure of the radius, or breadth, of the conceptual space activated during the conversation. We also measured conversational breadth by how far a conversation moved in the embedding conceptual space from the first few utterances of the conversation to the last few utterances of the conversation. Second, we measured the average conceptual distance traveled in each conversational turn by averaging the conceptual distances between the first and second utterances, the second and third utterances, and so on until the last utterance in the conversation. We also considered the relationship of these findings on many more languages with a simple conversational simulation described in the SI.

Finally, we asked whether these conversational patterns would be related to the knowledge output of social collectives that use the language in their communication (step 5). If a social collective uses an informationally denser language to communicate, then their collective communication would likely traverse the conceptual space in a more dense, narrow pattern as expected above, leading to collective outputs resembling the communication process that generated them. We test this expectation by measuring the conceptual breadth in over 90,000 Wikipedia articles authored in 140 languages.

In what follows, using large-scale computation, artificial intelligence techniques, and massive, parallel corpora, we here show substantial variation in the information density of languages (step 1) and that this variation is related to their semantic density (step 2), highlighting consequences for human communication and coordination. We demonstrate that higher density languages communicate information more quickly relative to lower density languages (step 3). Then, using over 5,800 real-life conversations across 14 languages (step 4) and 90,000 Wikipedia articles across 140 languages (step 5), we show that topics are discussed more narrowly and deeply in denser languages, with conversations and articles retracing and cycling over a narrower conceptual terrain. These results demonstrate an important source of variation across the human communicative channel, suggesting that the structure of language shapes the nature and texture of conversation, with important consequences for the behavior of groups, organizations, markets, and societies. Table 1 summarizes the theoretical and methodological framework of our study, highlighting the different analytical steps, their rationale, the dataset used to test each step, and the study's findings.

## Results

**1. Variation in information density.** Consistent with limited work that has estimated the information density of languages using modest language samples and small textual corpora[38,39], using a large-scale sample of 998 languages representing 101 language families and a broad diversity of knowledge domains we find substantial variation in linguistic information density across the world's languages (Fig. 1b). We further show consistency in language information density rates across diverse knowledge domains. We find that information density is highly correlated across knowledge domains as diverse as religion, banking, medicine, government, computer programming, news, movies, and public speeches (SI Fig. 1). This suggests that language information density is domain-independent and applies to all observed knowledge areas within each language. Information density represents variation distinct from previously measured linguistic variation within language. It posts correlations of $\rho=.11$ with a composite measure of complexity that codes each language as simple if reliant on a lexical strategy with few grammatical distinctions and complex if on many[65,66] (available for 580 of our languages), $\rho=.24$ with a measure of fusion that captures the degree to which a language relies on phonologically bound markers such as prefixes and suffixes instead of independent ones[67], and $\rho=-.04$ with a measure of informativity, which assess the amount of explicit and obligatory distinctions a language makes, including those of politeness and remoteness[67] (both available for 380 of our languages). We include these measures as controls in models that follow.

**2. Language information density and the semantic density of conceptual information**. We find broad variation in the global structure of conceptual space across languages (SI Fig. 2). Some languages more densely interlink diverse conceptual subspaces while others are sparser and more fragmented. This variation is strongly associated with language information density. Figure 2 displays raw associations between language information density and density in conceptual space across corpora, showing strong associations in all cases. We test a mixed effects regression with observations nested within languages and languages nested within language families while controlling for the number of words in the corpus and other attributes of the corpus and language, we find that a one unit increase in information density is associated with a 1.02 unit increase in conceptual space density (95% CI: 0.84, 1.2; *p*=0.00000e+00; SI Table 5). The association remains after we control for the morphological complexity, fusion, and informativity of the languages as detailed above (*β*=.98, 95% CI: 0.88, 1.08; *p*=0.00000e+00; SI Table 7). Thus, as more conceptual information is encoded into words of a language, concepts within that language become more closely associated as suggested in Fig. 1a.

While prior research has measured information density manually at the syllable or morpheme level,[39,68] it is impractical across many languages and massive corpora like ours, which prohibit manual coding. To ameliorate the concern that words might not be the most fundamental units of meaning and in line with prior work tracing information theoretic measures of language[69–71] we also created our language density measure at the single-character level. This yields similar results between information density and conceptual space density, with a coefficient of .61 (95% CI: 0.46, 0.74; *p*=2.22045e-16; Model 3 in SI Table 6). The robustness of these results to an alternative specification of information density suggests consistency across word and subword measures of information density.

**3. Language information density and communicative speed.** In contrast to prior research that found limited association between language information density and communication speed based on modest text in fewer than 20 languages[38,39], we find large variation in the speed of communication across languages and this speed is associated with information density (see Fig. 3). Using a random intercepts model, we find that a one unit increase in information density is associated with a .84 unit decrease in duration, which equates to an increase in communicative speed (95% CI: -1.19, -0.49; *p*=2.58014e-06; SI Table 12). We also tested the same model but with the character-level information density measure, finding consistent results (*β*=-0.77, 95% CI: -0.97, -0.57; *p*=1.64313e-14; SI Table 13). We further test a model with controls for the morphological complexity, fusion, and informativity of the language, again with consistent findings (*β*=-1.02, 95% CI: -1.35, -0.68; *p*=2.40372e-09; SI Table 14).

These findings are especially interesting when understood in the context of speech production and comprehension. During the process of human communication, speech encoding (i.e., articulation, or the process of turning information into speech sound) is the slowest part of speech production and comprehension, meaning that humans can think of what to say and understand what somebody else has said faster than it takes to say it[72]. In this way, there exists a processing time asymmetry between the mechanical work of speech articulation and the cognitive work of comprehension and inference. This asymmetry has been dubbed the articulatory bottleneck[72] and suggests that if inference-making is cheap (in time required) and articulation is expensive, then any increase in articulatory speed should likewise speed-up inference-making. It further suggests that the proportion of communication reliant on ampliative pragmatic inference will vary by language depending on information density, with higher density languages requiring more inference about what is spoken and why.

**4. Language information density and semantic breadth in real-world conversations.** We find evidence that informationally denser languages cover a *narrower* conceptual range across all conversations. Controlling for duration and the number of turns taken, conversations in informationally denser languages cover a conceptually narrower range of discussion (*β*=-0.053; 95% CI: -0.08, -0.03; *p*=.0001; SI Table 22; Fig. 4a). This effect is robust to the inclusion of controls for morphological complexity (*β*=-0.058; 95% CI: -0.09, -0.03; *p*=.00006; SI Table 23), environment including population size, total precipitation, and mean temperature (*β*=-0.054; 95% CI: -0.08, -0.03; *p*=.0002; SI Table 24), cultural attributes including indulgence and long-term orientation (*β*=-0.058; 95% CI: -0.08, -0.04; *p*=2.29792e-09; SI Table 25). These effects are both significant and substantial. Increasing informational density by one standard deviation (10) yields an increase in more than two-thirds (.675) of a standard deviation of conversational breadth. When we run an OLS model predicting conversational breadth with all controls described, the $R^2$ equals .55 (SI Table 27). Introducing information density to this specification increases the $R^2$ to .92 (SI Table 27), adding considerable explanatory power to the model.

These results indicate that the greater the information density of the language spoken in these conversations, the more conceptually narrow the conversations were across their entire content.

We next examined whether the conversations were also narrower in terms of their starting and ending locations in conceptual space. This goes beyond the previous measure to understand not just the total breadth of the conversation but also the movement between its conceptual starting and ending point. We find the same pattern, wherein conversations in informationally denser languages depart less from the initial topic of discussion by the end of the conversation. This effect was robust to how we operationalized the beginning and ending of the conversation, with similar results regardless of whether we measured the conceptual distance between the first 4 and last 4 utterances ($\beta$=-0.023; 95% CI: -0.04, -0.008; $p$=.002; SI Table 29), first 8 and last 8 ($\beta$=-0.012; 95% CI: -0.021, -0.004; $p$=.005; SI Table 30), utterances 2-6 and the last 6 through the last 2 utterances ($\beta$=-0.02; 95% CI: -0.03, -0.007; $p$=.002; SI Table 31), utterances 4-8 and the last 8 through the last 4 utterances ($\beta$=-0.02; 95% CI: -0.03, -0.006; $p$=.003; SI Table 32), and utterances 6-10 and the last 10 through the last 6 utterances ($\beta$=-0.02; 95% CI: -0.03, -0.005; $p$=.004; SI Table 33). In all of these cases, the conversations in informationally denser languages conceptually departed to a lesser degree from the initial topic of conversation.

The above results led us to conceive of a language generation model in which conversational discourse topics are sticky, with individuals tending to remain within a topic rather than moving away from it. We hypothesize that within denser languages, participants will be better enabled and more likely to discuss the same topic from different points of view and by bringing in more diverse information to bear on it. From this perspective, conceptually denser languages facilitate the mobilization of neighboring conceptual information, leading to a more thorough discussion of any given topic. When this occurs, conversations will be less likely to move to new topics, leading to narrower, but deeper, conversations overall. To explore this we test whether conversations in informationally denser languages activate and engage more deeply with the topics of discussion, staying to a greater degree within the conversational topics. For conversations, we trace the average conceptual distance moved between the first and last utterances, averaging the conceptual distance between the first and second utterances, the second and third utterances, and so forth until the final utterance. We find evidence that information density is related to narrow movements, with conversations in informationally denser languages engaging more deeply with each topic of conversation ($\beta$=-0.04; 95% CI: -0.06, -0.02; $p$=.00007; SI Table 28; Fig. 4b). These findings corroborate with results from simple conversational simulations estimated for many more languages (see SI Supplementary Text and SI Table 43).

## 5. Language information density and the semantic breadth of knowledge output from a social collective.

Finally, we find that collective knowledge outputs reflect the same conceptual dynamics found within the conversations that produced them. We trace the conceptual breadth of over 95,000 Wikipedia articles collectively written by hundreds of thousands of contributors in 140 languages. When we investigate whether the final articles written through online conversations exhibit different patterns of information content based on the language used to write them, we find support for the idea that informationally denser languages are associated with knowledge articles on Wikipedia that cover narrower conceptual terrain ($\beta$=-0.01; 95% CI:- 0.01, -0.006; $p$=0.004; SI Table 36; Fig. 4.c). These results are robust to the inclusion of language structure controls for morphological complexity, fusion, and informativity ($\beta$=-0.009; 95% CI: -0.02, -0.003; $p$=0.004; SI Table 37). Increasing informational density by one standard deviation (10.5) yields an increase in more than two-thirds (.673) of a standard deviation of discursive breadth. When we run an OLS model predicting article conceptual breadth with all of the controls, the $R^2$ equals .56 (SI Table 27). Upon introducing information density to this specification increases the $R^2$ to .71 (SI Table 27), significantly increasing the explanatory power of the model.

## Discussion

In summary, we report broad variation in how human languages encode information. While some languages more densely weave together diverse conceptual terrain, drawing together broad areas of conceptual space, others are sparser and more fragmented, separating distinct areas of meaning. We show that this variation in conceptual encoding is associated with important conversational dynamics, including speed of communication and patterned movement through the conceptual terrain of conversation and exposition. Because informationally denser languages contain more conceptual information per bit, more information flows through a fixed bandwidth communication channel leading to faster communication. Higher information density languages are also associated with conversations that intensively traverse a narrower portion of conceptual space. The set of possible things to say about any one topic is larger in denser languages, and opportunities for cycling are more numerous, leading conversation participants speaking these languages to trace and retrace the same conceptual terrain. Finally, we show that these conversational patterns are observable in expository knowledge outputs produced by collaborative groups, such that knowledge produced in higher information density languages more intensively covers a more narrow region of conceptual space.

Our study has focused on mapping the variation in information density across the world's languages and tracing its relationship with communication and conversation dynamics (the second arrow in Fig. 5). Nevertheless, our study raises the question of what drove observed differences in information density (the first arrow in Fig. 5). The first possibility is that languages will have evolved differing information density values simply by chance. Such a proposal would be supported by a "phono-semantic monkey" conception of language evolution.[73] The short version of this perspective is that the monkey's random typing on a keyboard would engender Zipfian word distributions without recourse to any kind of functionalist explanation. According to this argument, observed language information density variation is random, with languages randomly distributed within the "Goldilocks zone" permitted by human cognitive capacity and communicative need[74]. One point of evidence in support of this argument is that within our data we observe languages spoken within societies at advanced levels of cultural and economic development located both above and below the mean value within our data. In addition, we observe that languages spoken within simpler, less developed societies are also located at the lower and upper ends of the distribution, suggesting that there do not appear to be measured evolutionary or functionalist forces pushing toward more or less information density.

The second possibility, however, is that within the "Goldilocks zone"[74] environmental and social forces have shaped the amount of information density within languages. For example, prior work has documented how the number of language users, geographic spread, and degree of language contact relate to aspects of morphological complexity[65,66,75]. Evidence from our data for this position includes the much narrower position of the world's most widely spoken languages within the middle of the information density distribution, suggesting that modern culture and society has evolved a more optimal information density. This observation suggests that as the number of concepts required for encoding in language increases, the amount of information to be encoded per language unit approaches an optimum within the center of the distribution. Exploring mechanisms that might account for this narrowing of information density among the most widely spoken languages is an open avenue for future work.

In order to investigate the possibility that broader environmental factors shape language information density, we have added environmental measures[65,66] to our data, including size of the speaker population, number of neighbors, spatial perimeter of the language, mean temperature of the language's climate, yearly precipitation, and length of the growing season. When we test a random-intercepts model predicting information density, with observations nested within languages and languages nested within language families, we find that the land area ($\beta$=3.2; 95% CI: .4, 5.9; $p$=.02; SI Table 42), land perimeter ($\beta$=-9; 95% CI: -16, -2; $p$=.009; SI Table 42) and mean temperature ($\beta$=-1.3; 95% CI: -1.8, -.77;

*p*=1.84479e-06; SI Table 42) manifest statistical significance. Further, the map in Fig. 1c suggests regional differences in the variation of information density, with some regions converging toward a narrow band of values (e.g., southern India and southeast Africa) and others showing much broader variation (e.g., southern Mexico and Central America). These regional differences along with broader environmental forces at play in language evolution highlight the potential for future research to more deeply investigate the potential social and environmental forces at play in shaping language information density.

Our findings also suggest that because language information density is associated with patterns of communication, it might non-trivially shape individual cognition (e.g., creativity, memory, search), other unexamined patterns of interaction, and collective performance among human groups. In cognition, one can imagine that varying density of conceptual information, which shapes associational probabilities among language concepts, may influence many cognitive behaviors such as how individuals seek out novel information for problem-solving, judgment, and evaluation; how they store and activate conceptual memories; and how they engage in creative tasks[76]. In terms of social, variation in language information density may shape patterns of conflict and collaboration by altering the mobilization of concepts and ideas for social engagement. To the extent that the words and ideas of others during conversation function as cues to retrieve one's responses, differences in encoded information density could reshape interactive information activation and social encounters. Considering our demonstrated relationship between language structure and conversational speed and breadth, language structure may play a causal role in shaping patterns of social interaction and collective performance, broadening the linguistic relativity hypothesis beyond cognition. We hope this study can inspire investigations that extend it to social interaction and collective performance in collective knowledge, intelligence[77], and the evolution of culture, technology, and the built environment.

**Limitations.** While our observational empirical design provides support across a large number of languages for the claim that language information density relates to processes of social interaction and collective cognition, it does not allow us to make causal claims regarding this relationship. To establish causation, future research would benefit from two kinds of studies. Laboratory studies across smaller sets of languages can provide control over important conversational factors that stand to nudge patterns of interaction through language. For example, the McGrath task circumplex identifies tasks groups can engage in, ranging from cognitive tasks that require greater judgment and creativity to executive behaviors that require greater planning and coordination[78]. Across the broad range of tasks in which groups engage, some will require greater support for open collaboration while others the capacity to manage conflict. Experimental studies that standardize the nature of the task will enable deeper insight into whether and how language information density shapes patterns of communication and social interaction. By contrast, studies that engage in deeper ethnographic study of smaller language sets located on different ends of the information density continuum could provide important insights regarding how language information density relates to the pragmatic, cultural, and socio-structural factors that shape communication and social interaction in daily life. Both laboratory and ethnographic studies would enrich our understanding of processes through which language structure may shape communication, social interaction, and human performance across the many domains of collective life.

Second, the current study is agnostic about whether subdomains of conceptual meaning within and across languages might vary in terms of information density. While the current study suggests that information density tends toward a consistent level across different knowledge domains within a language (see the large correlations coefficients in SI Fig. 1 across different corpora), it is possible that at fine-grained levels, some subdomains of a language could be substantially denser than to others. For example, poetry and sociology are likely denser than logic and physics[79,80]. Understanding whether subdomain density is consistent across languages or whether there are substantial differences across which subdomains densely encode would better shape expectations for how social interaction and knowledge creation take place across language cultures.

**Conclusion.** Our findings call for an expansion of the linguistic relativity hypothesis beyond the cognitive framework, bringing the idea into the realm of communication, interaction, collaboration, and collective action. The structure of conceptual information within language not only shapes individual cognition [7,11,16,18,81,82], but how humans communicate and interact with one another, influencing the space of what collectives think and how they behave[29]. From coordination, cooperation, and collaboration to conflict, competition and disruptive innovation, the structure of information in language may impact the character, success and failure of human collectives by weaving the texture of their communication.

# Methods

No statistical methods were used to predetermine sample size. Randomization and blinding were not possible, given the observational nature of the study.

**1. Estimating language information density.**
*Dataset of parallel translations.* The following parallel translation corpora were collected in order to estimate the linguistic information and semantic densities. (1) complete Bible, including the Old and New Testaments[83–85]; (2) New Testament[83–85]; (3) news transcripts (News9 and News11)[83]; (4) web text[83,86]; (5) movie subtitles (Subs16 and Subs18)[83,87]; (6) TED talks[83]; (7) example sentences for foreign language learners; (8) United Nations[83,88]; (9) European Central Bank[83]; (10) European Medicines Agency[83]; (11) European Union bookshop[83]; (12) European Union Directorate General for Translation[83]; (13) European Union Joint Research Centre[83]; (14) European Parliamentary Proceedings[83]; (15) GNOME software files[83]; and (16) KDE software files[83]. SI Table 1 documents the number of languages and language families represented in each corpus. In total, these corpora represent 998 languages within 101 language families.

*Measuring language information density.* To estimate language information density, we draw on tools from information theory, which have been increasingly used in the study of language and meaning.[41,42,89] We use the text of each corpus above and count how many times each word appears. We then use this distribution to generate the most efficient binary code for words in each document using the Huffman coding algorithm[44]. This algorithm creates the most efficient prefix-free binary code with which to encode a given symbol distribution. Formally, the input to the algorithm is a symbol set $A = \{a_1, a_2, ..., n\}$ of size $n$, with weights $W = \{w_1, w_2, ..., n\}$, where $w_i = weight(a_i)$, $i \in \{1, 2, ..., n\}$ and is the proportion of each $a_i$ in symbol set $A$. In the case of language, the symbols can be characters, morphemes, or words. Because our interest lies in how conceptual information is encoded, we use the symbol distribution of the unique words within the transcript. The output of the algorithm is a code $C(W) = \{c_1, c_2, ..., n\}$, where $c_i$ is the binary codeword for $a_i$. We let

$$L(C(W)) = \sum_{i=1}^{n} w_i \times length(c_i)$$

be the weighted path length of code $C$, and satisfy the condition $L(C(W)) \leq L(T(W))$ for any code $T(W)$. The output of the Huffman Coding algorithm, then, is the shortest possible binary code for a given symbol distribution.

When every word in a text is translated into its Huffman code, we are left with a document of 0's and 1's and can count the number of binary digits (bits) in each Huffman coded document. For a given document, the larger the bit size of a document, the less informationally dense a language is, as this means that each bit contains less conceptual information. The smaller the bit size, the more informationally dense a language is, as this means that more conceptual information is contained within each bit. To make the measure comparable across corpora, we take the English document as the baseline bit size for each dyadic translation within a corpus and every other language takes on a ratio value relative to English. That is, the English value is the denominator and the ratio of each language is arrived at by using its document bit size in the numerator. We created this measure for 998 languages for which parallel translation corpora existed. For ease of interpretation of the statistical tests, we take this ratio and multiply it times -100, leading to larger values being associated with informationally denser languages. Figure 1b displays the distribution of languages across the information density continuum. It is worth noting that this approach is closely related to other information theoretic approaches to studying how information is distributed within the words of a language.[42]

**2. Language information density and the semantic density of conceptual information.**
*Measuring language semantic density.* To measure the density of the conceptual space, we created a 300-dimensional vector-space model of the text within each document used to construct the information density measure, using a continuous bag-of-words approach[47,90,91] and the following parameters: epochs=10 (number of iterations over the corpus), window = 5 (number of words before and after the focal word). These kinds of models project the word co-occurrences within a text into a multi-dimensional vector space wherein similar syntactic and semantic words tend to be close to each other and wherein words can have multiple degrees of similarity. Because these models can capture multiple degrees of similarity, they provide a useful representation of the conceptual space of a language as captured in any given corpus, given the multidimensional nature of word meaning. Vector-space models such as this effectively represent the conceptual relationships between the words of a language, capturing semantic regularities such as: V-king – V-man + V-woman = ~V-queen. The expectation is that languages with informationally denser words will tend toward conceptual spaces that are likewise denser, as each word within the space is likely to have on average more associational possibilities to every other word. The associational possibilities in higher information density languages are partly a consequence of words that appear in multiple contexts and that help to bring together disparate locations within the multidimensional spaces of these models.

Once the vector-space model for each individual document has been learned, we measured the average cosine similarity (which ranges between -1 and 1) between ten thousand random word-pairs for words within the vocabulary of the text. Larger values indicate closer distances. This cosine similarity measure characterizes the average distance within the multidimensional conceptual space of each document. A larger cosine similarity indicates that concepts tend to share closer meaning associations to each other, with a more compressed space between concepts. A smaller cosine similarity, on the other hand, indicates that more space has to be travelled in order to reach any given conceptual location. As with the linguistic information density measure, to make the measure comparable across corpora, we take the English document as the baseline density for each dyadic translation within a corpus and every other language takes on a ratio value relative to English. For ease of interpretation, we multiply the ensuing value by 100. Larger values equal denser conceptual spaces.

*Language family control.* To control for language family in our models, we use data from glottolog, a comprehensive catalog of the world's languages, language families, and dialects[92].

*Morphosyntactic complexity, fusion, and informativity controls.* To account for relevant typological features we mobilize three previously used measures. The first is a measure of morphosyntactic complexity, which uses 27 morphosyntactic variables and codes each strategy as either simple (if it relied on a lexical strategy or few grammatical distinctions) or complex (if it relied on a morphological strategy or many grammatical distinctions) [65,66]. The value of the measure is then the sum of the number of complex strategies within each language. This measure was available for 528 of our 1,390 observations. The second is a measure of fusion, which measures the degree to which a language relies on phonologically bound markers such as prefixes and suffixes instead of phonologically independent markers [67]. Languages with more phonologically fused markers lead to larger values of fusion. The third is a measure of informativity, which measures the amount of explicit and obligatory distinctions that a language makes, including features such as politeness distinctions in pronouns and remoteness distinctions in past and future tenses[67]. The greater the number of distinctions, the larger the value. The second and third measures were available for 380 of our 1,390 observations.

*Statistical analysis*. In this study, we are observing language information density estimates that are nested within languages, which are themselves nested within language families, leading to the potential that each observation will not be independent. Because it is possible that attributes of the corpora available within each language differentially influence the relationship, we fit a three-level mixed model with random intercepts at both the language family and language-within-language-family level. We estimated this model using Stata's mixed command, with standard errors estimated using the vce(robust) option[93], which uses the Huber-White sandwich estimators[94,95]. The complete specification can be found in the supplementary materials. We followed the same modeling strategy for steps 4 and 5 below.

### 3. Language information density and communicative speed.

*Dataset and measure of audio file duration times.* To measure the communicative speed of a language, we collected the duration time of the audio Bible spoken in different languages. The audio Bible duration data was manually collected from metadata of the MP3 files found at wordproject.org and faithcomesbyhearing.com. We have audio duration data for 265 languages representing 51 language families. As with the measures above, we again take the ratio of these duration times relative to English to make them comparable across websites. Higher values equal longer duration times and thus slower communicative speed.

A potential drawback with this measure is that we have only one observation per language and therefore cannot account for the distribution of individual speaking styles within a language. Nevertheless, we believe this weakness is overcome by the fact that we are able to estimate communicative efficiency at greater scale across a large sample of languages and for samples of spoken text that are orders of magnitude longer than has been done before[39,68]. This helps to address the shortcomings of prior measures, which relied on much smaller samples of spoken text (a few sentences) in only a handful of languages.

*Statistical analysis*. Because we only have one value of communicative speed for each language, we compute the mean value of information density for all measures within a language. For this model, language observations are nested within language families. Our estimated model then has random intercepts at the language family level. We estimated this model using Stata's mixed command, with standard errors estimated using the vce(robust) option[93], which uses the Huber-White sandwich estimators[94,95].

## 4. Language information density and semantic breadth in real-world conversations.

*Dataset of conversations.* Our conversational data came from the Babel Program of the Intelligence Advanced Research Projects Agency. These natural conversations are between individuals from broad age ranges, include speakers from both genders, and take place in a variety of environments including the street, home, office, public places, and vehicles. We used conversations from the following 14 languages: Amharic[96], Bengali[97], Cebuano[98], Georgian[99], Guarani[100], Igbo[101], Kazack[102], Lithuanian[103], Tagalog[104], Tamil[105], Telugu[106], Turkish[107], Vietnamese[108], and Zulu[109]. These languages represent eight language families. The mean number of conversational turns across all conversations is 78 with a standard deviation of 26.

*Measuring conceptual breadth of a conversation.* For each set of conversations within a language, we measured the conversational breadth of every conversation as follows. We took the transcripts of the conversation, converted the text into a list of words, and from this list we measured the centroid vector of the unique words within the list. We then measured the cosine similarity of each unique word to the centroid vector using the fasttext embedding models pre-trained on the Wikipedia corpus of the language[110]. We took the mean cosine similarity (times -1, so that larger values equal greater breadth) between all unique words and the centroid as an indication of the unstandardized conceptual breadth of the conversation. To standardize by the density of the conceptual space itself, we took the ratio of the conceptual breadth of the conversation over the density of the spaces trained on our parallel corpora.

*Measuring conversational distance of the conversation.* To measure the conceptual distance covered during the conversation, we followed a similar procedure to measuring the conceptual breadth, but instead of measuring the cosine similarity of unique words to the centroid vector we measured the distance between the first four utterances of a conversation and the last four utterances. We used the same Wikipedia embeddings as above[111]. For robustness we created the same measure for the first and last eight utterances. As with the conversational breadth measure, we likewise standardized these by the conceptual density of the language itself. For additional robustness, we measure the second through sixth, fourth through eighth, and sixth through tenth conversational turns at the beginning and end of the conversation, meaning that the very first and very last turns of conversation are not taken into account. The consequence of these additional measures is to move us away from the potential of capturing mere conversational formalities such as hello's and goodbye's and toward capturing the actual topics of conversation.

*Measuring average distance traveled.* We measured the average conceptual distance moved during a conversational turn by measuring the cosine similarity between the first and second utterances, the second and third utterances, and so on until the final utterance. As before, we normalized this measure by taking the ratio of this cosine similarity over the density of the conceptual space of the language. Figure 4b displays this distribution of average distance moved by conversational turn within the conversations of each language.

*Environmental controls.* To control for the potential effect of environmental variables that might influence patterns of social interaction, we control for population size, precipitation, and mean yearly temperature[65,66,112].

*Culture controls*. To control for the potential effect of cultural dimensions on patterns of communication and social interaction, we added controls for indulgence (which existed for 13 out of the 14 languages) and long-term orientation (which existed for 12 out of the 14 languages)[113,114]. Other cultural variables such as power distance, uncertainty avoidance, and masculinity were not available for many of the languages in our conversation corpora, so we did not include them in the analysis.

*Statistical analysis*. We followed a similar statistical modeling strategy here as we did for step 2 above. In this study, we are observing conversations that are nested within languages, leading us to observe multiple observations of conversations within a language. We therefore fit a two-level mixed model with random intercepts at the language-family level. We estimated this model using Stata's mixed command, with standard errors estimated using the vce(robust) option[93], which uses the Huber-White sandwich estimators[94,95]. The complete specification and associated output can be found in the supplementary materials.

**5. Language information density and the semantic breadth of knowledge output from a social collective.**

*Measuring article breadth of Wikipedia articles.* To measure the breadth of Wikipedia articles, we followed the following procedure. For each available pre-trained embedding model provided by Facebook's fasttext team[110] and available in our corpus of parallel translations (158 languages in total), we used the Wikipedia Python API to import and read the text of 500 random articles in that language. For each of these articles, we split the text into a set of words and we measured the cosine similarity of 100 random word pairs. We used the average cosine similarity of these 100 word pairs (times -1, so that larger values equal greater breadth) as a measure of the conceptual breadth of the article. As before, we also normalized this measure by the density of the conceptual space of that language.

*Statistical analysis*. We followed the same statistical modeling strategy here as we did for step 2 above, fitting a random intercepts model with Wikipedia articles nested within languages and languages nested within language families.

## Data Availability
The datasets analyzed within this study are publicly available from:
https://opus.nlpl.eu/
https://www.ldc.upenn.edu/
https://wordproject.org
https://faithcomesbyhearing.com
https://glottolog.org/
https://pypi.org/project/Wikipedia-API/

## Code Availability
The code used in this research was written in Python 3.7.2, with models run in Stata 17 and will be made available in a public repository upon publication.

## Author contributions

P.A. and J.A.E. performed the research design. P.A. analyzed the data. P.A. and J.A.E. did the writing.

**Competing interests**
The authors declare no competing interests.

# Tables

Table 1 | Theoretical and methodological framework for studying the relationship between language information density and communication

| Step | Rationale | Dataset | Finding |
|---|---|---|---|
| 1. Examine variation in information density (Huffman encoding of words) across the world's languages. | Establish a substantial empirical distribution of information encoding density across languages. | Parallel corpus covering 15 knowledge domains across 998 languages. | Broad variation in information density across the world's languages (Fig. 1b). |
| 2. Explore the relationship between language information density and the semantic density of conceptual information. | Provide evidence that languages which encode more conceptual information within their words create closer associations across the space of conceptual information. | Parallel corpus covering 15 knowledge domains across 998 languages. | Language information density is positively associated with semantic density (Fig. 2). |
| 3. Investigate the relationship between language information density and communicative speed. | Provide evidence of an expected information theoretic relationship between information density and the speed with which communication travels through a channel. | Audio Bible in 265 languages. | Language information density is positively associated with communicative speed (Fig. 3). |
| 4. Consider the relationship between language information density and semantic breadth in real-world conversations. | Provide evidence that the encoding of conceptual information evidenced in steps 1 and 2 correlates with patterns of communication in real-world conversations. | More than 9,000 real-world conversations in 14 languages. Conversational simulation in 82 languages. | Conversations in informationally denser languages cover a narrower conceptual range within a conversation (Fig. 4a-b). |
| 5. Consider the relationship between language information density and the semantic breadth of knowledge output from a social collective. | Provide evidence that the encoding of conceptual information evidenced in steps 1 and 2 correlates with patterns of real-world knowledge creation and communication. | Over 95,000 Wikipedia articles in 140 languages. | Collectives speaking informationally denser languages produce conceptually narrower articles (Fig. 4c). |

# Figures

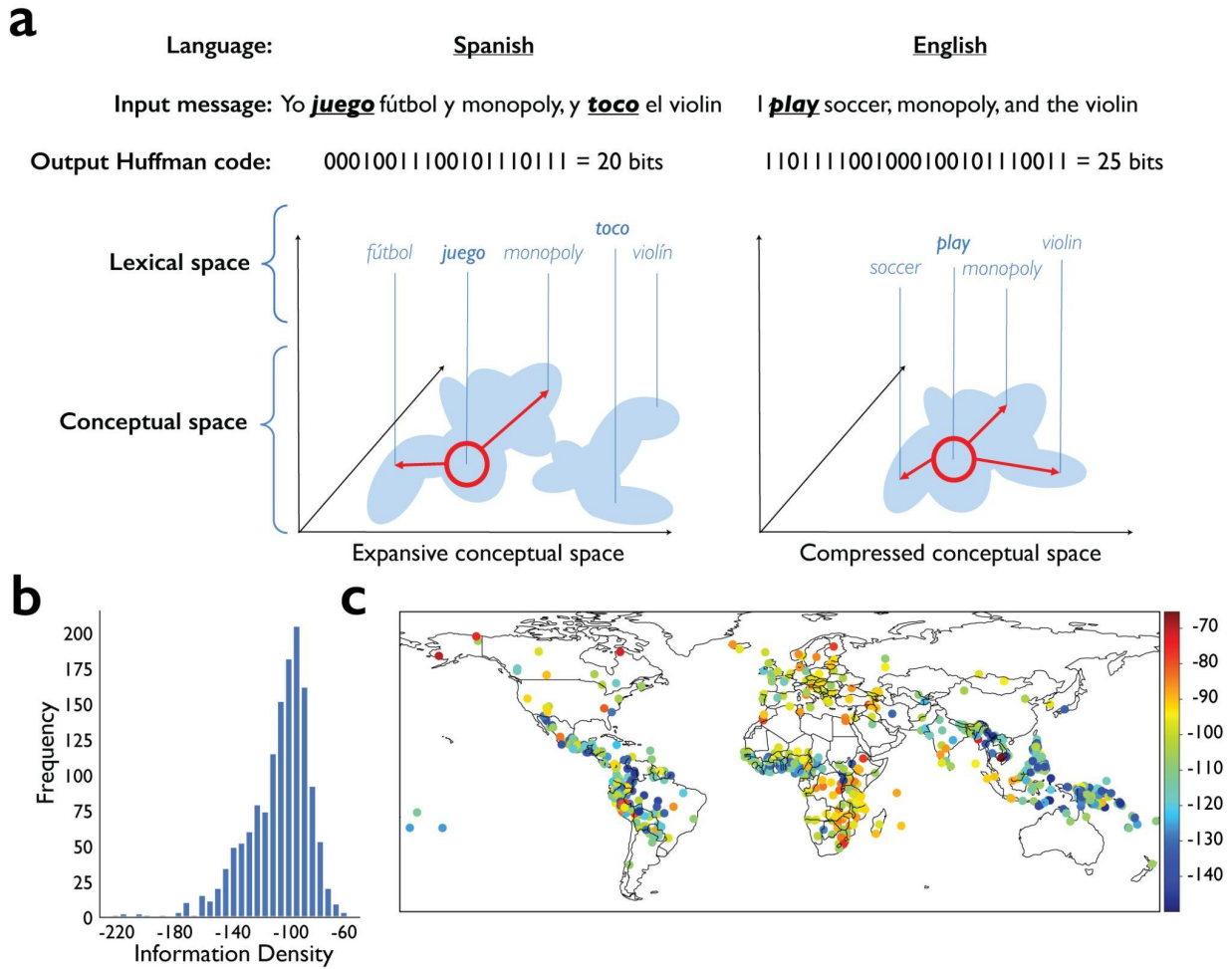

**Fig. 1 | Broad variation in the information density of human languages.**
**a**, Simplified illustration of information density, semantic density, and how they are related. Information density is estimated by taking parallel translations of an input message (i.e., the documents within our parallel translation corpora) and using the distribution of words within the document to compute an optimal, prefix free binary code for the words. We then translate the message into binary digits using this code. Informationally denser languages will have messages with fewer bits. Then, by taking the bit size ratio between the messages (times -100 so denser languages have larger values), we arrive at a measure of language information density. The consequence of informationally denser languages is that they compress the conceptual space. In the example above, because the word "play" in English is representing more conceptual information by also referring to the activity of moving a bow across the strings of a violin, the conceptual information of "monopoly" and "soccer" get nudged closer to the conceptual information related to "violin." Our information density measure carries out this procedure at scale across massive, parallel aligned corpora, giving us an estimate of the information density across the entirety of a language.
**b**, Distribution of the language information density measure. Larger values equal informationally denser languages. **c**, Geographic distribution of linguistic information density. Larger values (toward the red end of the continuum) equal informationally denser languages. Some regional clustering occurs (e.g., Europe, Southeast Africa, Southeast Asia).

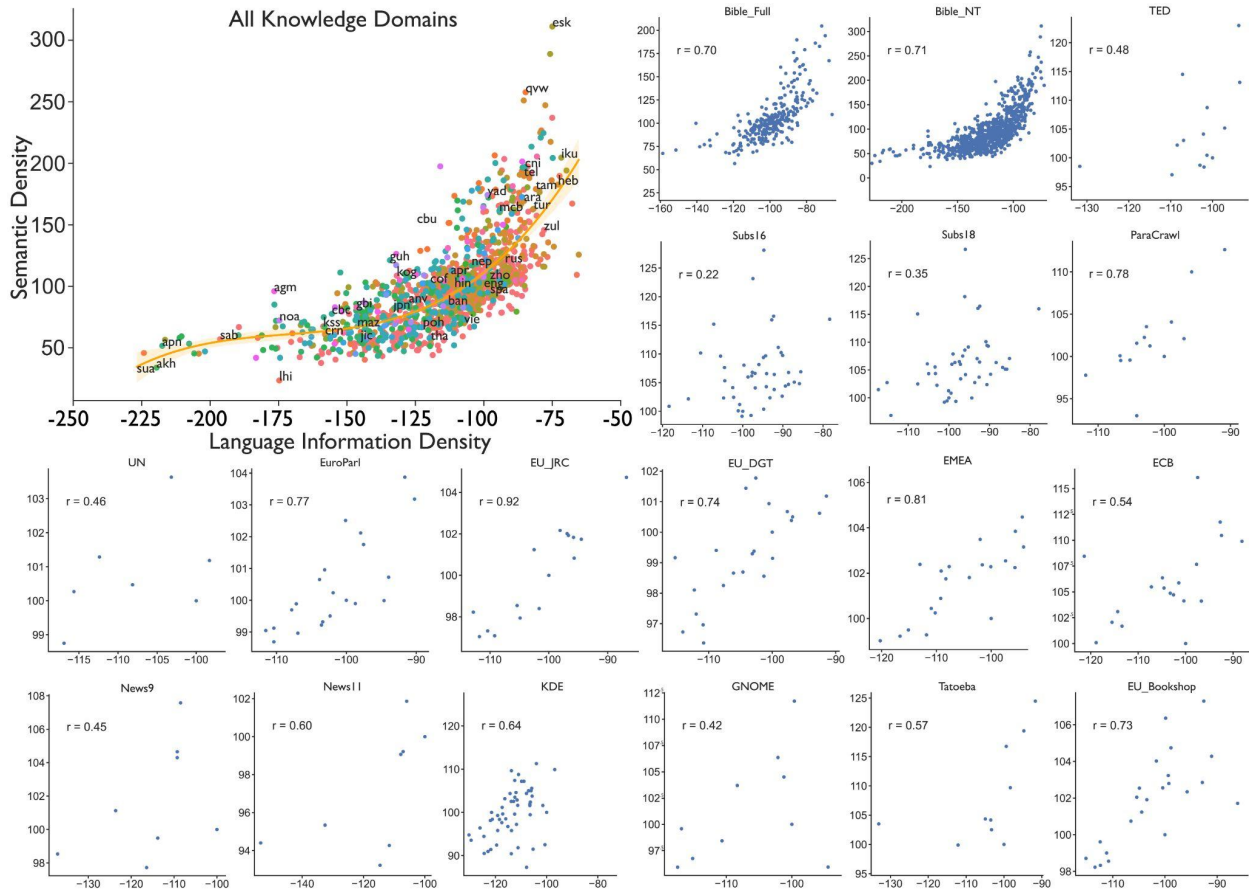

**Fig. 2 | Relation between information density and semantic density by knowledge domain.**
Across the 18 corpora in our sample, informationally more dense languages have conceptual or semantic spaces that are likewise more dense. For all plots, the *y*-axis marks semantic density while the *x*-axis indicates language information density. Language codes in Methods Table 1. The upper left plot includes the observations of all documents and includes the quadratic function with 95% confidence intervals represented by the shaded region.

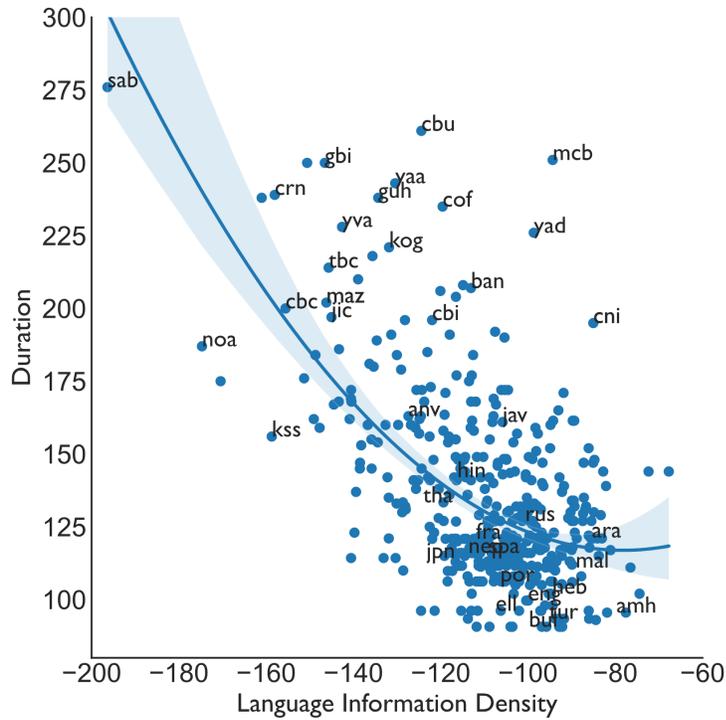

**Fig. 3 | Association between information density and communicative speed.**
The Y-axis is measured as duration, so smaller values equal greater communicative speed. Thus, informationally denser languages that pack information into fewer bits communicate more quickly. Language codes in Methods, Table 1. The shaded region represents 95% confidence intervals of the quadratic function.

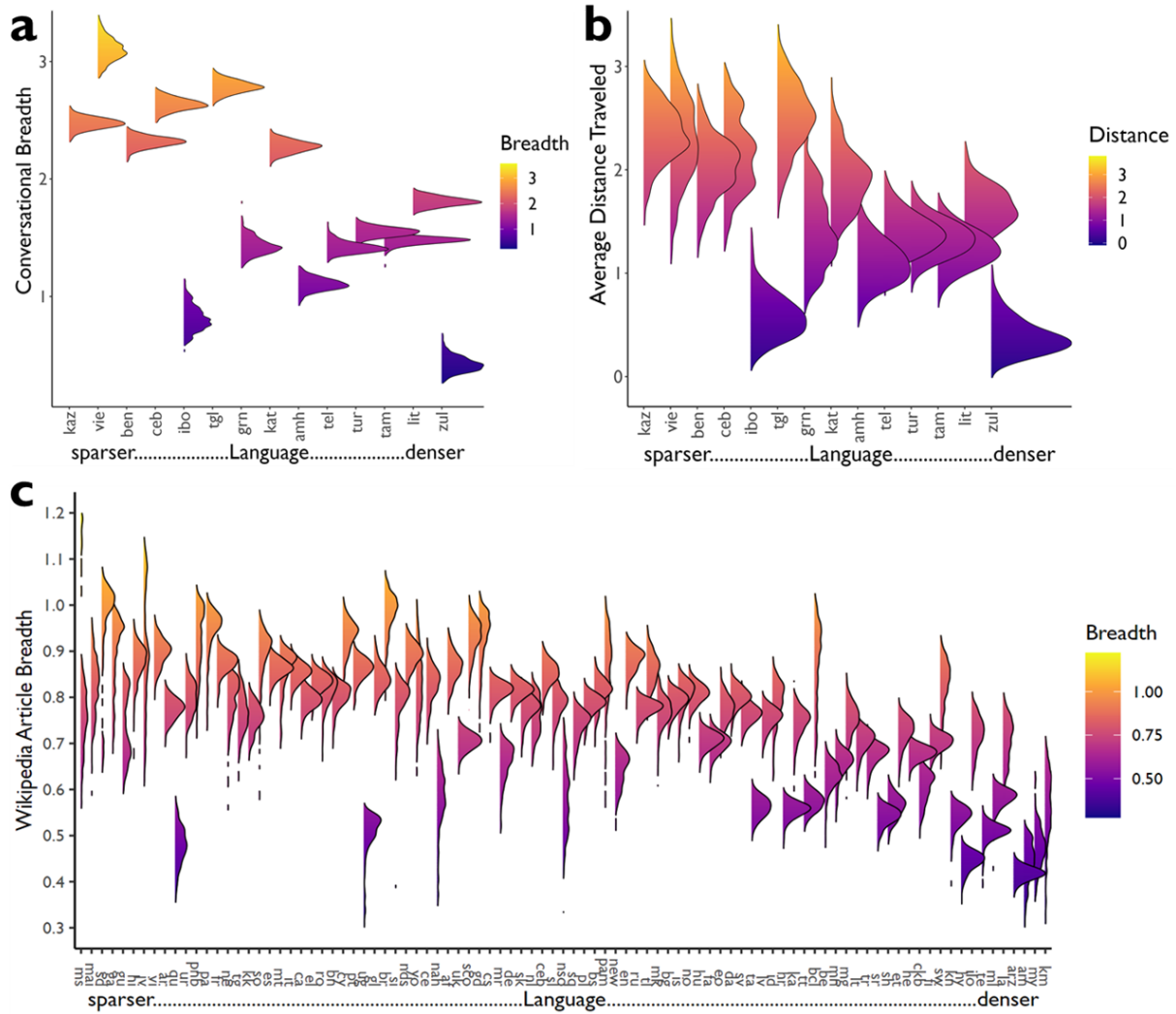

**Fig. 4 | Effect of information density on conversational patterns and knowledge output. a**, Association between information density and the conceptual breadth of conversations, with density estimates represented with vertical ridgelines to capture the modal pattern of association (censored below a cutoff of 1% of the maximum density). **b**, Association between information density and the average conceptual distance traveled during each conversational turn, with density estimates represented with vertical ridgelines to capture the modal pattern of association (censored below a cutoff of 1% of the maximum density). **c**, Association between information density and the conceptual breadth of Wikipedia articles, also plotted with density estimates represented with vertical ridgelines. As with the decreased breadth of conversations in 4a, the knowledge output of conversations in informationally denser languages are likewise narrower in conceptual breadth.

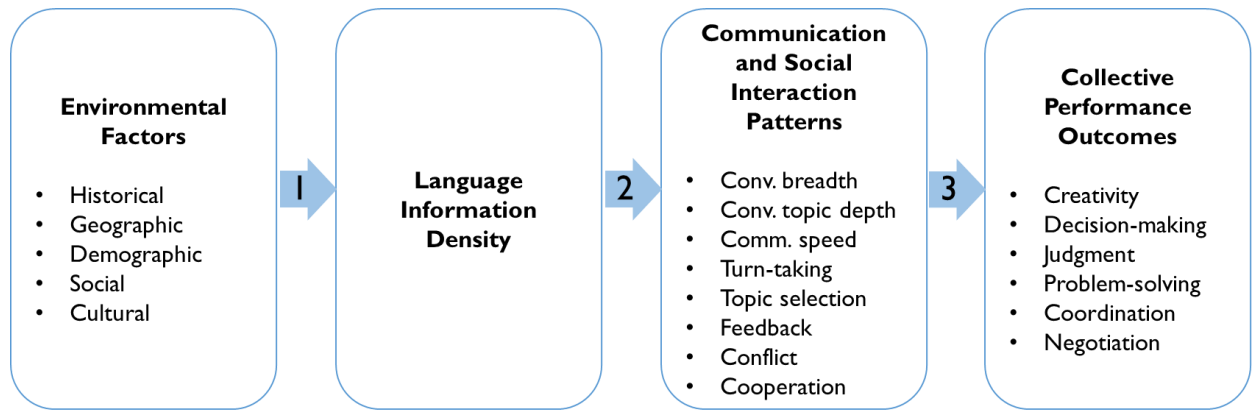

**Fig. 5 | Antecedents and consequences of language information density.**